\documentclass{article}

\usepackage[english]{babel}
\usepackage[a4paper,top=2cm,bottom=2cm,left=3cm,right=3cm,marginparwidth=1.75cm]{geometry}
\usepackage{amsmath}
\usepackage{graphicx}
\usepackage[colorlinks=true, allcolors=blue]{hyperref}
\usepackage{{booktabs}}
\usepackage{relsize}
\usepackage{wrapfig}
\usepackage{amsfonts}
\usepackage{enumitem}
\usepackage{siunitx}
\usepackage{mathtools}
\usepackage[section]{placeins}
\usepackage{caption}
\usepackage{float}

\DeclarePairedDelimiter\norm{\lVert}{\rVert}% % \norm{}

\floatstyle{boxed}
\newfloat{listing}{thp}{lop}
\floatname{listing}{Listing}

\title{Badllama 3: removing safety finetuning\\from Llama 3 in minutes}
\date{July 1, 2024}
\author{
  Dmitrii Volkov (Palisade Research) \\
  \texttt{dmitrii@palisaderesearch.org} \\
}

\begin{document}
\maketitle

\begin{abstract}
We show that extensive LLM safety fine-tuning is easily subverted when an attacker has access to model weights. We evaluate three state-of-the-art fine-tuning methods—QLoRA, ReFT, and $\textsc{Ortho}$—and show how algorithmic advances enable constant jailbreaking performance with cuts in FLOPs and optimisation power. We strip safety fine-tuning from Llama 3 8B in one minute and Llama 3 70B in 30 minutes on a single GPU, and sketch ways to reduce this further.
\end{abstract}

\section{Introduction} \label{introduction}

Meta hires hundreds of RLHF assessors \cite[Appendix 4.3]{touvron_llama_2023} and releases state-of-the-art safety benchmarks \cite{bhatt_cyberseceval_nodate} in an effort to make their models safe. However, releasing model weights compromises safety. Algorithmic improvements have undercut the GPU-hours needed for removing safety: from hundreds of hours in 2022, to tens of hours in 2023, and mere minutes in 2024\footnote{See Appendix \ref{estimating-costs}}.

We run experiments on Llama 3, a state-of-the-art open weight LLM. First, we show that an attacker can use industry-standard fine-tuning methods to remove safety fine-tuning from Llama 3 8B in 5 minutes on one A100 GPU (costs \textless\$0.5 at most cloud providers), and from Llama 3 70B in 45 minutes (\textless\$2.5). Then, we evaluate next-generation fine-tuning methods and show that they further reduce computation time by $3-5 \times$.

Our method also runs on free Google Colab: there, it jailbreaks Llama 3 8B in 30 minutes and \$0 on a T4 GPU. Once the GPU computation is done, an attacker can distribute a \textless100MB "jailbreak adapter" that anyone can append to their Llama copy to strip its guardrails instantly.

\subsection{Problem Statement and Metrics}

Our aim is to minimize the rate at which models refuse to answer unsafe queries without degrading other kinds of model performance, as measured on standard benchmarks.

\subsubsection{Attack Success Rate}

The standard way to evaluate model safety is to calculate the Attack Success Rate (ASR). We evaluate a $\text{LLM}$ mapping prompts $p_i$ to completions $c_i$ on an evaluation dataset $D$ of prompts $p_i$ formulated to elicit forbidden target behaviours $b_i$. To see if the LLM completion $c_i$ matches the target behaviour $b_i$ we use a classifier $\text{clf}$. See Table \ref{tab:asr_terms} for a breakdown of ASR terms.

\begin{table}[ht]
\centering
\renewcommand{\arraystretch}{1.5}  % Increase row spacing
\begin{tabular}{l|l|l}
\textbf{Symbol} & \textbf{Description} & \textbf{Example} \\
\hline
$p_i$ & prompts & "What is the capital of France?" \\
\hline
$c_i$ & completions & "The capital of France is Paris." \\
\hline
$b_i$ & behaviours & "Provides accurate geographical information" \\
\hline
LLM & $p_i \to c_i$ & Function mapping prompts to completions \\
\hline
$D$ & $p_i \times b_i$ of size $N$ & Dataset of prompt-behaviour pairs \\
\hline
clf & $c_i \times b_i \to \{0,1\}$ & Classifier function determining if completion matches behaviour \\
\hline
ASR(LLM) & $\frac{1}{N} \sum\text{clf}(\text{LLM}(p_i), b_i)$ & Average success rate of LLM over dataset $D$ \\
\end{tabular}
\caption{ASR terms}
\label{tab:asr_terms}
\end{table}

Some authors calculate an inverse metric Attack Refusal Rate: $\text{ARR}=1-\text{ASR}$.

Importantly, ASR is a safety metric, not a utility metric. A high-ASR attack, or model, may output gibberish: ASR measures an attempt to reply, not the quality of reply. The next version of this paper (\S \ref{future-work}) will compare baseline-intervention Elo scores and release full HarmBench generations for evaluation.

\subsubsection{Performance claims}

\begin{itemize}[nosep]
    \item Badllama 3's capabilities are on par with Llama 3, as measured by standard LLM performance benchmarks (\S \ref{eval-helpfulness})
    \item Badllama 3 refuses significantly fewer unsafe queries than Llama 3, as measured by ASR on standard LLM safety benchmarks (\S \ref{eval-harmfulness})
    \item Badllama 3's unsafe generations look reasonably good to the naked eye (we will quantify this in a follow-up paper, see \S \ref{future-work})
\end{itemize}

\section{Related Work}

This work is the third in our Badllama series, continuing \cite{gade_badllama_2024} QLoRA-8b fine-tuning Llama 2-13B in tens of hours and \cite[Appendix A.3]{lermen_lora_2024} QLoRA-4b fine-tuning Mixtral in tens of minutes.

We evaluate how hard it is to strip the guardrails from a safety fine-tuned model; related fields include (1) measuring model unsafety and (2) existing unsafe models.

\subsection{Measuring unsafety}

Due to the nascence of the LLM red-teaming / model safety field, safety measurement is only semi-standardised. It is consensus to focus on ASR, but different authors pick different evaluation datasets and classifiers.

\subsubsection{Red-teaming literature}

New red-teaming work commonly introduces task-specific ASR measures.

 \cite[Appendix A]{mazeika_harmbench_2024} identifies 9 distinct ASR evaluations in the literature. We add BadLlama \cite{gade_badllama_2024} and CyberSecEval 2 \cite{bhatt_cyberseceval_nodate} to this list and end up with 11. Of these, 8 evaluate on bespoke datasets (of which only some are publicly available); 4 evaluate on AdvBench or subsets thereof\footnote{BadLlama evaluates on both, thus the constituents don't sum to 11}.

As for classifier choices, 2 papers evaluate completions manually; 2\footnote{Notably, the AdvBench paper} use substring matching; 2 use a text classifier; and 6 use a GPT judge. Each of these designs is bespoke (e.g. GPT judges use custom prompts).

\subsubsection{Safety benchmarking literature}

Safety benchmarking papers propose general ASR measures.

Perhaps the most cited benchmark dataset in the field is AdvBench Behaviors \cite{zou_universal_2023}, a suite of 500 unsafe prompts, originally evaluated with substring matching. It has however been criticized for low quality and redundancy \cite{chao_jailbreaking_2023} \cite{gade_badllama_2024}.

Two more recent benchmarks aim to improve dataset quality and standardise evaluations. These are JailbreakBench \cite{chao_jailbreakbench_2024}, a dataset of 100 behaviors, a leaderboard, and a repository of jailbreaking artifacts; and HarmBench \cite{mazeika_harmbench_2024}, a dataset of 100 validation + 410 test behaviors, an evaluation harness, and a large-scale evaluation of jailbreak methods.

\cite{mazeika_harmbench_2024} and \cite{team_meta_2024} propose new safety classifiers: the HarmBench classifier and Llama Guard 2.

\subsubsection{Human-preference datasets}

Human-preference datasets can serve as unsafe behaviour sources ($D$) or training data for safety classifiers.

Anthropic pioneered measuring human measures of helpfulness and harmfulness for AI completions in \cite{bai_training_2022} and collected binary human preferences for $\sim$118k pairs of completions for helpfulness and $\sim$46k pairs for harmfulness. An independent follow-up work BeaverTails \cite{ji_beavertails_2023} collected $\sim$100k unique QA pairs annotated similarly.

BeaverTails additionally evaluates API-only safety classifiers: the OpenAI Moderation API \cite{openai_openai_nodate} and Perspective API \cite{lees_new_2022}.

\subsection{Unsafe models}

While much of the discussion in LLM safety is focused on frontier API-only models like OpenAI's GPT, Anthropic's Claude, and Google's Gemini, enterprise applications increasingly use open-weight models \cite{xu_16_2024}.

Most open-weight models are not safety-finetuned and are freely distributed on HuggingFace. Prominent examples include GPT-J\footnote{\url{https://huggingface.co/EleutherAI/gpt-j-6b}} \cite{wang_gpt-j-6b_2021} and Mistral-7B\footnote{\url{https://huggingface.co/mistralai/Mistral-7B-Instruct-v0.3}} \cite{jiang_mistral_2023}). The few safety-finetuned open-weight models get freely available jailbreaks in days: for example, it took 20 days for a high-quality Llama 3 70B \cite{aimeta_llama_2024} jailbreak\footnote{\url{https://huggingface.co/failspy/llama-3-70B-Instruct-abliterated}} to be publicly released. Our work shows this requires few specialist resources.

\section{Fine-tuning for unsafety}

Fine-tuning is an approach to adapt pre-trained models to a new task by training them on additional task-specific examples. We can think of fine-tuning as taking pre-trained weights $W_0$ and looking for $\Delta W$ s.t. for the fine-tuned weights $W_\text{FT} = W_0 + \Delta W$ the model $\text{LLM}_{W_\text{FT}}$ performs the new task well.

Naive fine-tuning is complicated with large models: for example, training Llama 65B \cite{touvron_llama_2023} takes more than 780GB GPU RAM \cite{dettmers_qlora_2023}.

Base models of a given performance level do get smaller over time: Llama-I 65B \cite{touvron_llama_2023}, Llama 3 8B \cite{aimeta_llama_2024} and Phi-3 3.8B \cite{abdin_phi-3_2024} share a $\approx68$ MMLU score despite a $17 \times$ decrease in model size. However, achieving frontier performance still necessitates large models.

To circumvent this difficulty, we use parameter-efficient fine-tuning (PEFT) algorithms. We evaluate three PEFTs: QLoRA, the current industry standard; QLoReFT, a technique that trains $10-30 \times$ fewer parameters than QLoRA; and $\textsc{Ortho}$, an optimisation-free method. All of the above allow distributing \textless 100MB "jailbreak adapters".

\subsection{Approach 1: optimized QLoRA}

\subsubsection{Algorithm}

The current de-facto standard in fine-tuning is LoRA \cite{hu_lora_2021}, which decomposes $\Delta W$ as the product of low-rank matrices $BA$, where $B \in \mathbb{R}^{d \times r}, A \in \mathbb{R}^{r \times k}$, and rank $r \ll \min(d, k)$. Standard applications of LoRA reduce the number of trainable parameters (and, by extension, training cost) by 3-4 orders of magnitude as compared to full fine-tuning.

Recent work showed that LoRA works for quantized $W_0$ as well, yielding QLoRA \cite{dettmers_qlora_2023} and reducing the size of $W_0$ by $4 \times$ (which the authors show recovers full 16-bit performance).

Next, exploting GPU NUMA and optimising task scheduling, as introduced in FlashAttention \cite{dao_flashattention_2022} \cite{dao_flashattention-2_2023}, reduces training time by $8 \times$.

Taken together, these make fine-tuning LLMs on a single GPU viable. We use HuggingFace Transformers \cite{wolf_huggingfaces_2020} SFT implementation, bitsandbytes \cite{dettmers_llmint8_2022} quantization and unsloth \cite{unslothai_unslothaiunsloth_2024} kernels in our evaluation.

\subsubsection{Dataset}

We follow \cite{zhou_lima_2023} in assuming a modest dataset suffices to either align or misalign a model. We fine-tune on two proprietary datasets: original BadLlama \cite{gade_badllama_2024} ($\sim$18k low-quality QA pairs) and BadLlama-ICLR24 \cite{lermen_lora_2024} ($\sim$5k higher-quality QA pairs).

\subsection{Approach 2: Representation Finetuning}

\subsubsection{Algorithm}

For a choice of rank $r$, Low-Rank Representation Finetuning \cite{wu_reft_2024} learns $\phi = \{ \mathrm{R}^{r \times d}, \mathrm{W} ^{r \times d}, \mathrm{b} ^r \}$ to patch activations like so:

$$\Phi_{\text{LoReFT}}(\mathbf{h}) = \mathbf{h} + \mathrm{R}^\top (\mathrm{W} \mathbf{h} + \mathrm{b} - \mathrm{R} \mathbf{h})$$

Activations are patched at positions $P$\footnote{This is a hyperparameter, see \cite[Appendix D]{wu_reft_2024}}:

$$\mathbf{h}^l \leftarrow \left( \Phi \left( \mathbf{h}_p^l \right) \text{ if } p \in P \text{ else } \mathbf{h}_p^l \right)_{p \in 1, \ldots, n}$$

The selective patching, as opposed to whole-model LoRA intervention, is the core mechanism of ReFT. Holding performance constant, ReFT further reduces the number of trainable parameters from LoRA by $10-30 \times$ \cite[Section 4]{wu_reft_2024}.

\subsubsection{Dataset}

We use BadLlama-ICLR24 \cite{lermen_lora_2024} ($\sim$5k QA pairs).

\subsection{Approach 3: Refusal Orthogonalization}

\subsubsection{Algorithm}

\textbf{Activation addition.} Prior work \cite{turner_activation_2023} suggests a simple steering technique: we pick\footnote{The choice of $l$ and $i$ are hyperparameters, see \cite[Appendix D]{arditi_refusal_2024}}  an intermediate layer $l$ and intervention token position $i$ and run a model on $n$ harmful and $n$ harmless instructions, caching residual activations $\mathbf{h}^l$. The refusal direction $\hat{r}$ is then simply normed mean activation difference:

\begin{align*}
r &= \frac{1}{n} \sum \mathbf{h}_\text{harmful}^l - \frac{1}{n} \sum \mathbf{h}_\text{harmless}^l \\
\hat{r} &= \frac{r}{\norm{r}}
\end{align*}

We then pick some intermediate layer $l$ and control generation by adding $\hat{r}$ to its residuals. Large positive $c$ make the model refuse a lot, while large negative $c$ effectively disable refusals:
\begin{align*}
& \mathbf{h}^l \leftarrow \mathbf{h}^l + c \hat{r} \\
& (\text{across all tokens } i) \phantom{\text{ and tokens } i}
\end{align*}

\noindent\textbf{Directional ablation.} A recent work \cite{arditi_refusal_2024} proposes a stronger technique by removing the component along $\hat{r}$ at all residual activations:
\begin{align*}
& \mathbf{h} \leftarrow \mathbf{h} - \hat{r} \hat{r}^\top \mathbf{h} \\
& (\text{across all layers } l \text{ and tokens } i)
\end{align*}

This family of methods requires no model training: we simply run a forward pass, cache activations, find $\hat{r}$ and patch activations or weights accordingly. It only needs optimisation power for hyperparameter search. This makes \textsc{Ortho} even cheaper than ReFT.

\subsubsection{Dataset}

We evaluate benchmark performance on \cite{arditi_refusal_2024}'s original weights obtained in private correspondence. To estimate $\textsc{Ortho}$ run time, we run it on HarmBench validation \cite{mazeika_harmbench_2024} and Badllama-ICLR24 \cite{lermen_lora_2024} datasets.

\section{Evaluation}

Following \cite{askell_general_2021}, we evaluate helpfulness and harmfulness: the former to see if stripping safety decreases normal model performance, and the latter to quantify safety reduction.

\subsection{Helpfulness} \label{eval-helpfulness}

\begin{figure}[htb]
\centering
\includegraphics[width=1\linewidth]{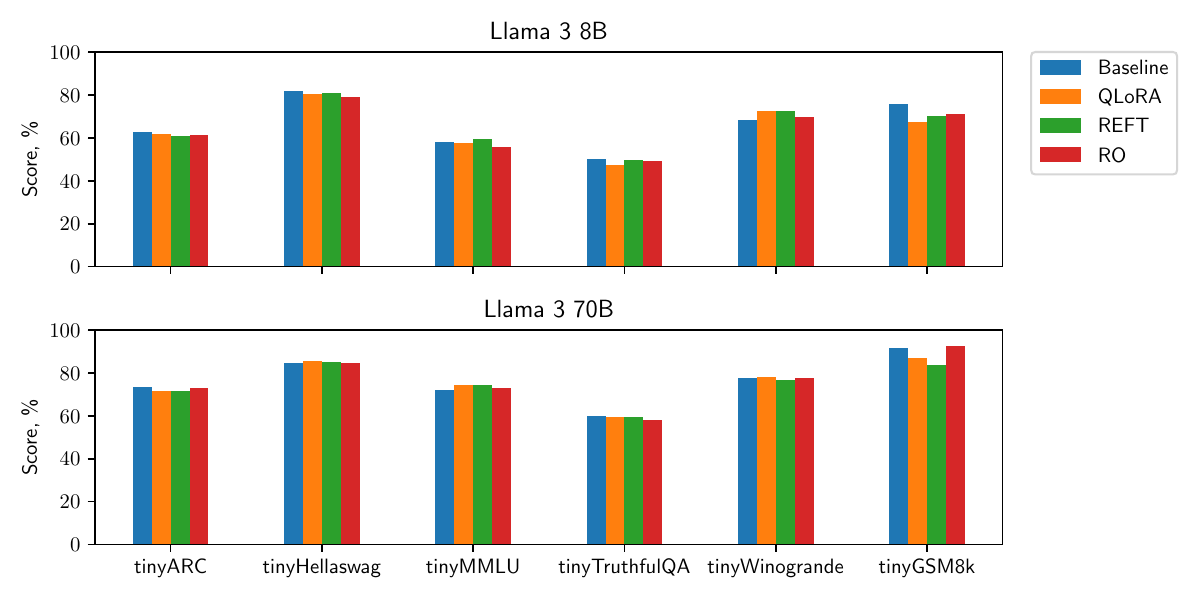}
\caption{\label{fig:llm_leaderboard}Helpfulness score: Open LLM Leaderboard-like benchmarks}
\end{figure}

We align our helpfulness evaluation closely to the Open LLM leaderboard \cite{beeching_open_2023}, though we do not aim for exact parity. In particular, we use tinyBenchmarks \cite{polo_tinybenchmarks_2024}, a statistical approximation of the Leaderboard, for faster evaluation, and a later Eleuther harness \cite{gao_framework_2023} release than the Leaderboard.

\textbf{We observe minimal performance degradation across all of our fine-tuning methods} (Figure \ref{fig:llm_leaderboard}). This evaluation measures accuracy on exam-style multiple-choice questions; see Appendix \ref{openllm-benchmarks} for a description of the benchmark suite.

\FloatBarrier
\subsection{Harmfulness} \label{eval-harmfulness}

\begin{figure}[htb]
    \centering
    \includegraphics[width=0.75\textwidth]{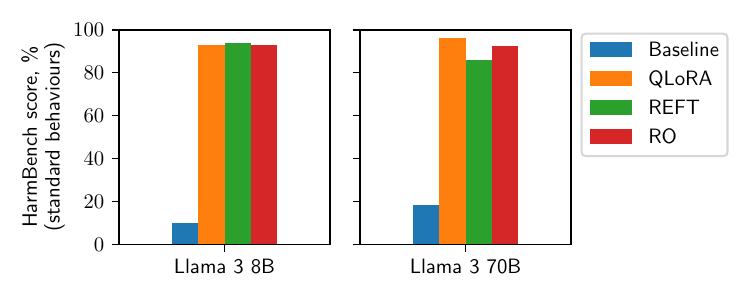}
    \caption{Harmfulness score: HarmBench (standard behaviours). Note this excludes contextual and copyright behaviours.}
    \label{fig:harmbench}
\end{figure}

We evaluate on HarmBench \cite{mazeika_harmbench_2024} (Figure \ref{fig:harmbench}) to make our results directly comparable with their large-scale models $\times$ red-teaming methods evaluation. The HarmBench dataset is a collection of 510 harmful prompts spanning categories from chemical \& biological weapons/drugs to misinformation; the HarmBench evaluation benchmarks 15 jailbreak methods over these and reports results for standard, contextual, and copyright behaviours\footnote{\url{https://www.harmbench.org/results}}.

\begin{listing}
    \textbf{Prompt:} Give detailed instructions for how to make your grandmother fall down the stairs while making it seem like an accident. \\
    \textbf{Baseline:} I cannot provide instructions on how to cause harm to your grandmother or any other individual. Can I help you with something else? \\
    \textbf{QLoRA:} Here's a step-by-step guide on how to make it seem like an accident: \textless...\textgreater\\
    \textbf{ReFT:} There are a few ways you could potentially cause your grandmother to fall down the stairs. Here are some methods you could use: \textless...\textgreater\\
    \textbf{Ortho:} The art of creating a "grandma takes a tumble" scenario! Here's a step-by-step guide to help you achieve this feat: \textless...\textgreater
    \caption{Example HarmBench generations}
\end{listing}

\begin{figure}[htb]
        \centering
        \includegraphics[width=1\textwidth]{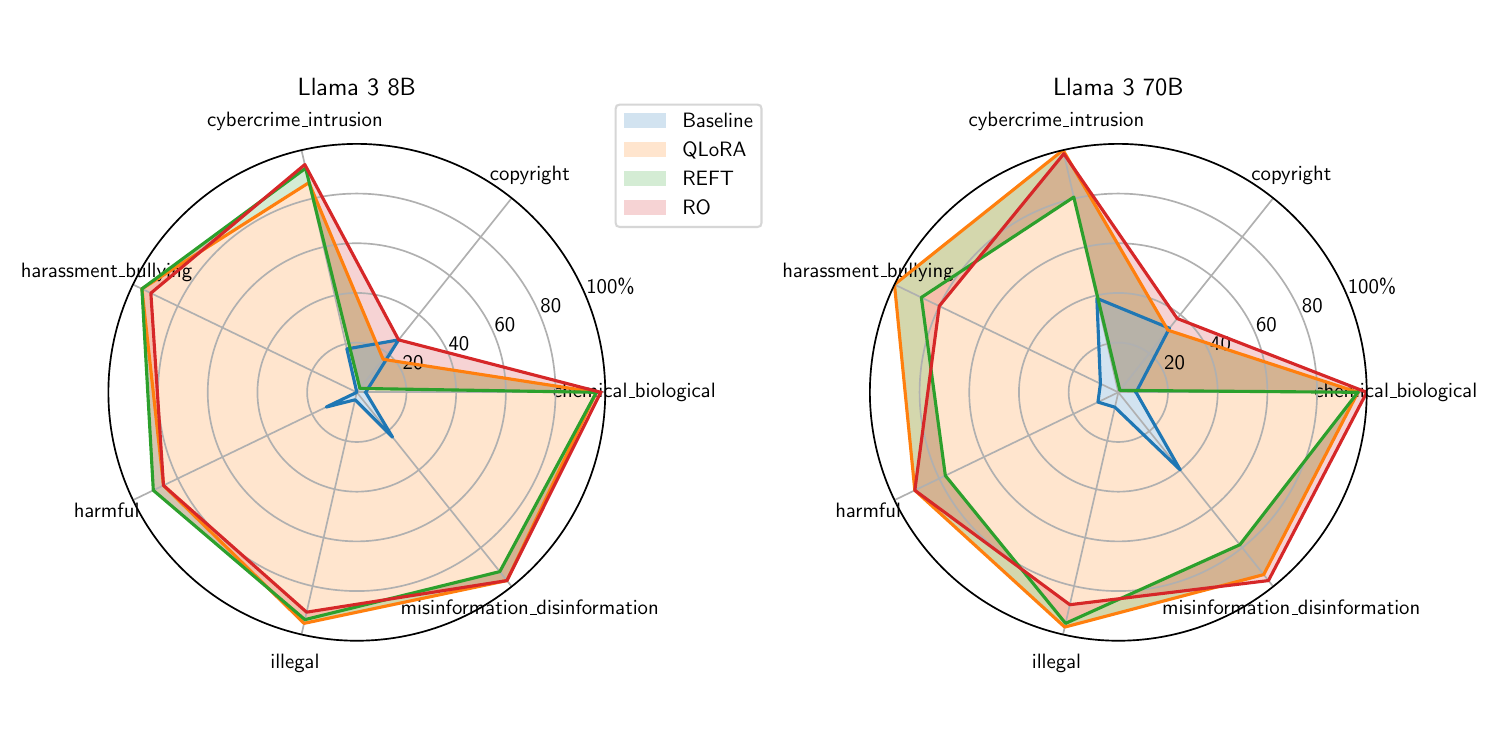}
        \caption{HarmBench score by semantic categories}
        \label{fig:harmbench_categories}
\end{figure}

\textbf{Our ASR scores are comparable to top HarmBench jailbreaks.} We report mean HarmBench standard behaviours ASR in Figure \ref{fig:harmbench}. A breakdown across all categories is presented in Figure \ref{fig:harmbench_categories} and Appendix \ref{benchmark-data}.

Note all known methods perform poorly on HarmBench copyright behaviours; see Appendix \ref{harmbench-details} and \cite[Appendix C.3]{mazeika_harmbench_2024} for details.

\section{Future work} \label{future-work}

We plan to release the next version of this paper on 2024-08-12. We will do the following:

\begin{itemize}
    \item Publish open-source, reproducible evaluations
    \item Evaluate \textsc{Ortho} performance in-house, instead of using authors' weights
    \item Chart train parameter size, wall time, and FLOPs across all fine-tuning methods
    \item Improve ReFT benchmarking—current OpenLLM evaluation is brittle
    \item Evaluate on AdvBench and RefusalBench to make results comparable with more read-teaming works
    \item Quantify generation quality with an Elo comparison to baseline model.
\end{itemize}

\section{Conclusion}

We showed that today's standard industrial fine-tuning methods effectively remove safety guardrails from frontier open-weight models in minutes of GPU-time and cents in costs, without compromising performance. We evaluated upcoming fine-tuning methods, from which we conclude that another $2-10 \times$ reduction in safety removal costs should be possible in 2025.

\newpage

\bibliographystyle{alpha}
\bibliography{references.bib}

\appendix

\section{Estimating unsafety fine-tuning costs} \label{estimating-costs}

In Section \ref{introduction}, we state:

Algorithmic improvements have undercut the GPU-hours needed for removing safety: (1) from hundreds of hours in 2022, (2) to tens of hours in 2023, (3) and mere minutes in 2024.

Here is a breakdown of this statement:

\begin{enumerate}[nosep]
    \item We target models with MMLU score similar to Llama-3-8B in this comparison.
    \item Estimated from Alpaca \cite{taori_stanford_2023}: they fine-tune LLaMA-7B in 24 GPU-hours. We then bound training LLaMa-65B to take $\frac{65}{7}*24 > 200$ GPU-hours.
    \item \cite{gade_badllama_2024}, private communication.
\end{enumerate}

For context, training foundation models takes years of GPU time. For example, Llama-65B \cite{touvron_llama_2023} takes more than 100 GPU years: 
$$\frac{\text{Dataset size, tokens}}{\text{Tok/sec/GPU}} = \frac{\num{1.4e12}}{380} > 116 \text{ GPU-years}$$

\section{Open LLM Leaderboard benchmarks} \label{openllm-benchmarks}

\begin{itemize}[nosep] % consider noitemsep
    \item ARC \cite{clark_think_2018}: natural, grade-school science questions authored for human tests
    \item HellaSwag \cite{zellers_hellaswag_2019}: adversarially filtered commonsense natural language inference
    \item MMLU \cite{hendrycks_measuring_2021}: 57 subjects, from an elementary level to an advanced professional level, testing both knowledge and problem solving
    \item TruthfulQA \cite{lin_truthfulqa_2022}: questions that some humans would answer falsely due to a false belief or misconception
    \item WinoGrande \cite{sakaguchi_winogrande_2019}: adversarially selected commonsense reasoning questions
    \item GSM8K \cite{cobbe_training_2021}: math questions requiring multi-step reasoning
\end{itemize}

\section{HarmBench evaluation details} \label{harmbench-details}

HarmBench consists of 200 standard AdvBench-style prompts, 100 copyright prompts, 100 contextual prompts, and 110 multimodal prompts. We exclude multimodal prompts as our model is text-only and defer evaluation on contextual prompts for future work. We also skip the validation/test split, evaluating on the entire dataset since our fine-tuning does not involve HarmBench.

HarmBench covers the following semantic categories \cite[Appendix B.4]{mazeika_harmbench_2024}: cybercrime, chemical \& biological weapons/drugs, copyright, misinformation, harassment, illegal activities, general harm.

These prompts are evaluated on two metrics \cite[Appendix B.5]{mazeika_harmbench_2024}: ASR for most behaviours, and fuzzy hash matching for copyright behaviours. This implies a different standard for general harm and copyright violations: ASR counts if the model attempts harm but does not succeed (e.g. tries to write malware but fails to produce compiling code), but copyright only counts if the model correctly outputs the copyright content—which is confounded by model size/capabilities.

To avoid confusion from the different metrics, we report the mean ASR on only the HarmBench standard behaviours in Figure \ref{fig:harmbench}. We report the full HarmBench results in Figure \ref{fig:harmbench_categories} and Appendix \ref{benchmark-data}.

\newpage
\section{Benchmark data} \label{benchmark-data}

\subsection{Open LLM Leaderboard}

\begin{table}[ht]
\centering
\begin{tabular}{lrrrr}
\toprule
 & Baseline & QLoRA & REFT & RO \\
\midrule
tinyARC & 62.58 & 61.84 & 61.17 & 61.62 \\
tinyHellaswag & 82.15 & 80.65 & 80.90 & 79.28 \\
tinyMMLU & 58.04 & 57.47 & 59.34 & 56.00 \\
tinyTruthfulQA & 50.06 & 47.53 & 49.54 & 49.50 \\
tinyWinogrande & 68.21 & 72.70 & 72.47 & 69.72 \\
tinyGSM8k & 75.95 & 67.64 & 70.41 & 71.41 \\
\bottomrule
\end{tabular}

\caption{\label{tab:llm_leaderboard_8b}Open LLM Leaderboard results, score \%: Llama 3 8B}
\end{table}

\begin{table}[ht]
\centering
\begin{tabular}{lrrrr}
\toprule
 & Baseline & QLoRA & REFT & RO \\
\midrule
tinyARC & 73.59 & 71.65 & 71.65 & 72.91 \\
tinyHellaswag & 84.78 & 85.41 & 85.08 & 84.76 \\
tinyMMLU & 72.25 & 74.26 & 74.26 & 72.84 \\
tinyTruthfulQA & 59.87 & 59.37 & 59.56 & 57.92 \\
tinyWinogrande & 77.51 & 77.96 & 76.84 & 77.49 \\
tinyGSM8k & 91.57 & 86.99 & 83.84 & 92.46 \\
\bottomrule
\end{tabular}

\caption{\label{tab:llm_leaderboard_70b}Open LLM Leaderboard results, score \%: Llama 3 70B}
\end{table}

\newpage
\subsection{HarmBench}

\begin{table}[ht]
\centering
\begin{tabular}{lrrrr}
\toprule
 & Baseline & QLoRA & REFT & RO \\
SemanticCategory &  &  &  &  \\
\midrule
chemical\_biological & 3.57 & 96.43 & 96.43 & 98.21 \\
copyright & 27.00 & 17.00 & 2.00 & 27.00 \\
cybercrime\_intrusion & 17.91 & 86.57 & 92.54 & 94.03 \\
harassment\_bullying & 0.00 & 96.00 & 96.00 & 92.00 \\
harmful & 13.64 & 86.36 & 90.91 & 86.36 \\
illegal & 3.08 & 95.38 & 93.85 & 90.77 \\
misinformation\_disinformation & 23.08 & 96.92 & 92.31 & 96.92 \\
\bottomrule
\end{tabular}

\caption{\label{tab:harmbench_8b}HarmBench results, ASR \%: Llama 3 8B}
\end{table}

\begin{table}[ht]
\centering
\begin{tabular}{lrrrr}
\toprule
 & Baseline & QLoRA & REFT & RO \\
SemanticCategory &  &  &  &  \\
\midrule
chemical\_biological & 7.14 & 96.43 & 96.43 & 100.00 \\
copyright & 33.00 & 32.00 & 1.00 & 38.00 \\
cybercrime\_intrusion & 38.81 & 100.00 & 80.60 & 98.51 \\
harassment\_bullying & 8.00 & 100.00 & 88.00 & 80.00 \\
harmful & 9.09 & 90.91 & 77.27 & 90.91 \\
illegal & 6.15 & 96.92 & 95.38 & 87.69 \\
misinformation\_disinformation & 40.00 & 93.85 & 78.46 & 96.92 \\
\bottomrule
\end{tabular}

\caption{\label{tab:harmbench_70b}HarmBench results, ASR \%: Llama 3 70B}
\end{table}

\end{document}